\documentclass[sigplan,screen]{acmart}
\usepackage{booktabs}
\usepackage{multirow}
\usepackage{graphicx}
\AtBeginDocument{%
  \providecommand\BibTeX{{%
    \normalfont B\kern-0.5em{\scshape i\kern-0.25em b}\kern-0.8em\TeX}}}

\setcopyright{acmlicensed}
\copyrightyear{2025}
\acmYear{2025}

\begin{document}
\newcommand{\avash}[1]{{\color{blue}{\small\bf\sf [Avash: #1]}}}

\newcommand{\yzy}[1]{{\color{magenta}{\small\bf\sf [Yu: #1]}}}

\title{Towards Transparent AI: A Survey on Explainable Large Language Models}



\author{
   Avash Palikhe$^{1}$, Zhenyu Yu$^{2}$, Zichong Wang$^{1}$, Wenbin Zhang$^{1}$\\
  $^{1}$Florida International University, Miami, Florida, USA \\
  $^{2}$Universiti Malaya, Kuala Lumpur, Malaysia \\
}





\begin{abstract}
Large Language Models (LLMs) have played a pivotal role in advancing Artificial Intelligence (AI). However, despite their achievements, LLMs often struggle to explain their decision-making processes, making them a 'black box' and presenting a substantial challenge to explainability. This lack of transparency poses a significant obstacle to the adoption of LLMs in high-stakes domain applications, where interpretability is particularly essential. To overcome these limitations, researchers have developed various explainable artificial intelligence (XAI) methods that provide human-interpretable explanations for LLMs. However, a systematic understanding of these methods remains limited. To address this gap, this survey provides a comprehensive review of explainability techniques by categorizing XAI methods based on the underlying transformer architectures of LLMs: encoder-only, decoder-only, and encoder-decoder models. Then these techniques are examined in terms of their evaluation for assessing explainability, and the survey further explores how these explanations are leveraged in practical applications. Finally, it discusses available resources, ongoing research challenges, and future directions, aiming to guide continued efforts toward developing transparent and responsible LLMs.
\end{abstract}

\maketitle



\section{Introduction}
In the rapidly advancing field of natural language processing, large language models (LLMs) like BERT\cite{devlin2018bert}, T5\cite{raffel2020exploring}, GPT-4\cite{openai2023gpt4}, and LLaMA-2\cite{touvron2023llama} have demonstrated impressive capabilities across domains such as machine translation, code generation, medical diagnosis, and personalized education~\cite{lee2020biobert,chen2021evaluating,ng2024educational,lopes2020litetrainingstrategiesportugueseenglish}. However, their 'black box' nature, stemming from vast parameters and extensive training data, obscures internal mechanisms and decision-making processes, complicating explainability. This lack of transparency can lead to unintended issues like model hallucinations and harmful content, posing significant challenges in high-stake fields like health, finance, and law, where interpretability is critical to avoid serious errors. Without insight into decision-making, user trust erodes, and ethical concerns escalate.

Researchers have developed various explainable artificial intelligence (XAI) methods to address concerns about LLMs, revealing their internal processes and decision-making mechanisms with clear, human-level explanations. This is crucial for building user trust, ensuring ethical high-stake decisions, and identifying issues like model hallucinations and biases (e.g., gender bias in translating 'doctor' to male and 'nurse' to female pronouns). XAI also aids in debugging by highlighting attention patterns or context errors, improving model performance and reliability for real-world use. However, current XAI literature lacks a systematic focus on LLMs, with many surveys covering traditional models broadly and few addressing the unique challenges of transformer-based architectures (encoder-decoder components with distinct attention mechanisms). A systematic survey is needed to standardize taxonomies and tailor solutions for these architectures' specific issues.

This survey offers a systematic review of explainable artificial intelligence (XAI) methods in large language models (LLMs), highlighting recent advances, evaluation mechanisms, explanation applications, and future research directions. To the best of our knowledge, this is the first comprehensive survey providing a standard taxonomy based on transformer architectures (encoder-only, decoder-only, and encoder-decoder models), addressing specific challenges in each, and exploring evaluation and application of these explanations. By comparing various XAI methods across architecture-aware perspectives, it delivers a holistic view of techniques and their use cases. Key \textbf{contributions} include: 1) A detailed review of XAI methods with a novel taxonomy, comparing techniques and use cases across LLM architectures; 2) Analysis of evaluation mechanisms and practical applications of XAI explanations; 3) Identification of available resources, research challenges, and future directions for transparent, ethically aligned models.

\section{Background}

\subsection{Large Language Models}

Large Language Models (LLMs) have become fundamental to modern natural language processing (NLP), driving progress in applications such as machine translation, dialogue generation, question answering, and code synthesis~\cite{brown2020language, openai2023gpt4}. These models are primarily built upon the Transformer architecture~\cite{vaswani2017attention}, which uses self-attention mechanisms to encode contextual relationships between tokens. Trained on massive text corpora, LLMs learn to capture statistical and semantic patterns in language~\cite{touvron2023llama}, enabling them to generalize across a wide variety of tasks. As the size of these models grows, in terms of both the number of parameters and the diversity of training data, they begin to exhibit emergent behaviors such as in-context learning, multi-task generalization, and the ability to follow natural language instructions~\cite{wei2022emergent}. 


Despite their impressive capabilities, LLMs remain largely opaque. Their internal decision-making processes are encoded in high-dimensional and deeply layered representations, making it difficult for users to understand how specific predictions are made~\cite{bommasani2021opportunities, zhao2023explainability}. This lack of interpretability presents risks in safety-critical domains such as healthcare, law, and education, where incorrect or biased outputs can have serious consequences. Therefore, improving the transparency and accountability of LLMs has become an urgent priority, motivating the development of explainability methods tailored to their unique architecture and behavior.

\subsection{Explainability in LLMs}

Explainability refers to the ability to make a model’s internal logic and outputs understandable to humans. This property is essential for ensuring transparency, detecting failure modes, and enabling responsible AI deployment in domains such as healthcare, law, and finance~\cite{rudin2019stop, zhao2023explainability}. In the context of large language models (LLMs), explainability is particularly challenging due to their black-box nature. These models rely on deeply stacked Transformer layers and encode knowledge in high-dimensional hidden states, which makes tracing how specific outputs are generated inherently difficult~\cite{yu2022interaction, mavrepis2024xai}.

To address these challenges, researchers have proposed various strategies aimed at increasing the interpretability of LLMs. Some methods focus on saliency-based attribution, which highlights the contribution of input tokens to the model's output~\cite{dang2025multimodal}, while others use probing techniques to examine the internal representations at different layers. More recent techniques such as Chain-of-Thought (CoT) prompting have demonstrated that prompting format can shape how intermediate reasoning steps are expressed~\cite{wei2022chain}. Despite these advances, many current methods are limited in scope, failing to generalize across tasks or model architectures.  Importantly, LLMs differ significantly in architecture and each type poses distinct interpretability challenges.

\section{Taxonomy}
\begin{figure}[!htb]
    \centering
     \includegraphics[width=0.45\textwidth]{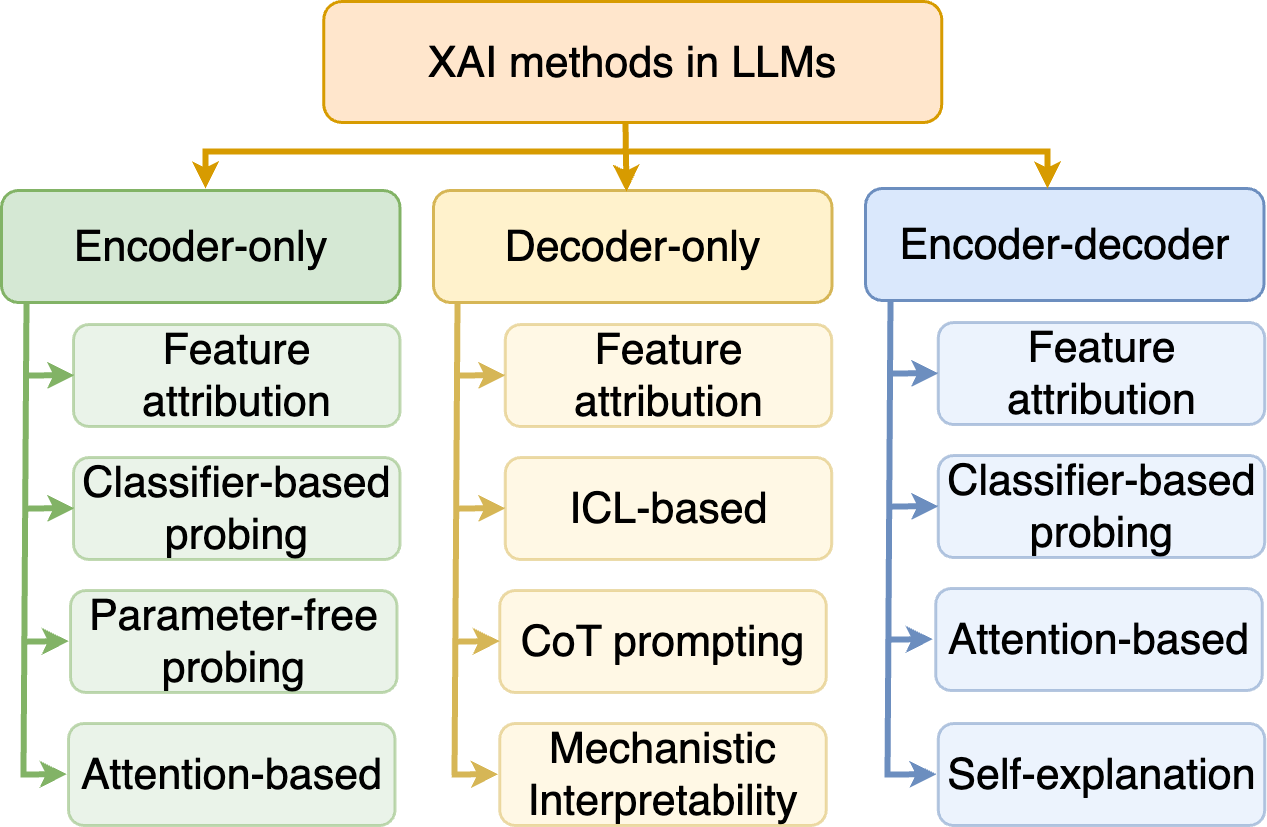}
    \vspace{+0.2cm}
    \caption{Taxonomy of XAI methods in LLMs.}
    \vspace{+0.2cm}
    \label{fig:taxonomy}
\end{figure}

We present a novel taxonomy for categorizing XAI methods based on the underlying transformer architecture of LLMs. This classification organizes the XAI approaches into three fundamental groups, as illustrated in Figure ~\ref{fig:taxonomy}. First,~\textit{XAI methods in encoder-only LLMs} are designed to interpret contextual embeddings, leveraging the bidirectional structure of the models to effectively analyze internal representations. Second,~\textit{XAI methods in decoder-only LLMs} rely on prompt engineering and in-context learning strategies to uncover the internal reasoning processes of these models, which are often opaque due to their massive scale and autoregressive nature. Third,~\textit{XAI methods in encoder-decoder LLMs} can provide insights at both the encoding and decoding stages, utilizing the dual architecture to interpret the flow of information through the cross-attention mechanism. This standardized taxonomy offers a comprehensive review of XAI techniques tailored to distinct architectures, enabling meaningful comparisons of their underlying structures and characteristics. It also addresses the unique challenges of achieving explainability in each type of architecture and highlights how these methods can be used for practical applications.

\begin{table*}[t]
  \vspace{+0.2cm}
  \caption{Summary of different methodologies used for XAI in LLMs}
  \label{tab:xai-methods}
  \centering
  \scriptsize
  \begin{tabular}{@{} l p{1.5cm} p{2cm} l p{2cm} l  p{1cm} l p{3.2cm}@{}}
    \toprule
    \textbf{Paper} & \textbf{Method} & \textbf{Type} & \textbf{Architecture} & \textbf{Model} & \textbf{Access} & \textbf{Scope} & \textbf{Integration} & \textbf{Application} \\
    \midrule
    ~\cite{szczepanski2021new} & LIME & Feature attribution-based & Encoder-only & BERT & External & Local & Post-hoc & Fake news detection classification \\
    \midrule
    ~\cite{lundberg2017shap} & TransSHAP & Feature attribution-based & Encoder-only & BERT & External & Local & Post-hoc & Tweet sentiment classification \\
       \midrule
    ~\cite{tenney2019you} & Edge probing & Classifier-based probing & Encoder-only & BERT & Internal & Local, Global & Post-hoc & NLP labeling tasks \\
       \midrule
    ~\cite{tenney2019you} & Token representation probing & Classifier-based probing & Encoder-only & BERT & Internal & Local & Post-hoc & Surface, syntax, and semantic probing tasks \\
        \midrule
    ~\cite{wu2020perturbed} & Perturbation masking & Parameter-free probing & Encoder-only & BERT & Internal & Global & Post-hoc & syntactic
parsing, discourse dependency parsing,  aspect based
sentiment classification \\
    \midrule
    ~\cite{wu2020perturbed} & DIRECTPROBE & Parameter-free probing & Encoder-only & BERT & Internal & Global & Post-hoc & Supersense-role, Supersense-function, POS, Dependency Relation, Semantic Relation \\
    \midrule
    ~\cite{vig2019analyzing} & BertViz & attention-based & Encoder-only & BERT & Internal & Local & Post-hoc & Detecting model bias, locating relevant
attention heads, linking neurons to model behaviour  \\
    \midrule
    ~\cite{hoover-etal-2020-exbert} & ExBERT & attention-based & Encoder-only & BERT & Internal & Local & Post-hoc &  Dependency relations, coreference, entity relationships
    \\
    \midrule
    ~\cite{barkan2024llm} & AML & Feature attribution-based & Decoder-only & GPT-2 & Internal & Local & Post-hoc & Faithful input attribution \\
    \midrule
    ~\cite{kariyappa2024progressive} & Progressive Inference & Feature attribution-based & Decoder-only & OPT, GPT-2 & External & Local & Post-hoc & Stepwise token attribution \\
    \midrule
    ~\cite{lampinen2023complementary} & Complementary Explanations & ICL-based analysis & Decoder-only & GPT-3 & External & Local & Prompt-time & Prompt structure analysis \\
    \midrule
    ~\cite{li2023saliencyicl} & Saliency-Guided ICL Analysis & ICL-based analysis & Decoder-only & GPT-2, GPT-Neo & External & Local & Post-hoc & ICL prompt token attribution \\
    \midrule
    ~\cite{zhang2024sea} & SEA-CoT & CoT prompting & Decoder-only & GPT-3.5 & External & Local & Prompt-time & CoT reasoning consistency \\
    \midrule
    ~\cite{zhou2022least} & Least-to-Most Prompting & CoT prompting & Decoder-only & GPT-3 & External & Local & Prompt-time & Hierarchical CoT reasoning \\
    \midrule
    ~\cite{meng2022locating} & ROME & Mechanistic interpretability & Decoder-only & GPT-2 & Internal & Local & Post-hoc & Factual memory editing \\
    \midrule
    ~\cite{gurnee2024language} & ACDC & Mechanistic interpretability & Decoder-only & GPT-2 & Internal & Local & Post-hoc & Abstract reasoning attribution \\
     \midrule    ~\cite{liu2024reliabilityexplainabilitylanguagemodels} & Program generation explanations & Feature attribution-based & Encoder-decoder & T5, CodeT5, CodeReviewer, CodeT5+ & External & Local & Post-hoc & Code repair, code review, code translation, and text-to-code generation \\
     \midrule
    ~\cite{enouen2024textgenshap} & TextGenSHAP & Feature attribution-based & Encoder-decoder & T5-large, T5-XXL and T5-FiD & External & Local & Post-hoc & Abstractive question answering from retrieval-augmented documents \\
     \midrule
    ~\cite{koto2021discourse} & Document-level discourse probing & Classifier-based probing & Encoder-decoder & T5, BART & Internal & Global & Post-hoc & Sentence ordering, discourse connective prediction, RST nuclearity prediction, RST relation prediction and cloze story test \\
     \midrule
    ~\cite{li2021implicit} & Implicit meaning representation probing & Classifier-based probing & Encoder-decoder & T5, BART & Internal & Global & Post-hoc & Next-instruction prediction, action
    and game-response generation task  \\
    \midrule
    ~\cite{juraska2021attention} & Attention-Guided Decoding & Attention-based & Encoder-decoder & T5 & Internal & Local & Post-hoc & Data-to-text NLG alignment \\
    \midrule
    ~\cite{cao2021attention} & Attention Head Masking & Attention-based & Encoder-decoder & BART, PEGASUS & Internal & Local & Post-hoc & Summarization content selection \\
    \midrule
    ~\cite{narang2020wt5} & WT5 & Self-explanation & Encoder-decoder & T5-small, T5-large & Internal & Local & Prompt-time & Answer justification generation \\
    \midrule
    ~\cite{yordanov2022few} & Few-Shot NLE Transfer & Self-explanation & Encoder-decoder & T5-base & External & Local & Prompt-time & Explanation transfer across domains \\
    \bottomrule
  \end{tabular}
  \vspace{0.5em}
\end{table*}

\section{Overview}

Based on the proposed taxonomy, this section offers an overview of explainable artificial intelligence (XAI) methodologies for large language models (LLMs), as detailed in Table~\ref{tab:xai-methods}. Studies are categorized across eight dimensions: Paper, Method, Type, Architecture, Model, Access, Scope, and Integration, with Application highlighting practical use cases. The Method (e.g., feature attribution-based, classifier-based probing, attention-based) and Type define the interpretability approach, while Architecture (encoder-only, decoder-only, encoder-decoder) and Model (e.g., BERT, GPT-2, T5) identify targeted LLM frameworks. Access (internal or external) and Scope (local or global) reflect data dependency and interpretability range, and Integration (post-hoc, prompt-time) indicates the implementation stage. Applications cover diverse tasks like fake news detection, sentiment classification, code repair, and abstractive question answering, showcasing the methods' ability to tackle specific interpretability challenges across LLM architectures.

\section{XAI methods in LLMs}
\label{sec:XAI_methods}

This section provides a detailed overview of XAI methods, categorized by three major model architectures, each presenting unique interpretability challenges and driving distinct methodological advances. The first category, encoder-only models, focuses on interpreting stable contextual embeddings from their bidirectional structure. The second, decoder-only models, targets autoregressive generation, often using prompt engineering to assess behavior. The third, encoder-decoder models, emphasizes interactions between encoder and decoder components, with a focus on cross-attention mechanisms. This categorization reveals shared methodological foundations and architecture-specific considerations influencing current and future XAI research.



\subsection{XAI methods in encoder-only LLMs}
XAI methods for encoder-only models explain contextual representations from their bidirectional attention mechanism\cite{devlin2018bert}, producing stable, semantically rich embeddings analyzed via techniques like feature attribution and classifier-based probing\cite{zhao2023explainability}. Masked language modeling (MLM) training simplifies explanation by allowing XAI to trace input token impacts on internal states or predictions, avoiding complexities of generative decoding or encoder-decoder interactions. Key methods include feature attribution, classifier-based probing, parameter-free probing, and attention-based approaches.




\sloppy
\subsubsection{Feature attribution-based methods}
\label{feature-attribution-encoder-only}

Feature attribution-based methods assess the relevance of features (e.g., words, phrases) to a model's prediction\cite{zhao2023explainability}, assigning scores reflecting each feature's contribution to the output. These techniques suit encoder-only models, pretrained with masked language modeling (MLM)\cite{devlin2018bert}, which integrates bidirectional left and right contexts, yielding stable, context-rich embeddings for analyzing feature influence on predictions~\cite{wu2021explainingexplanationsbertempirical}.

A widely used feature attribution-based XAI method, LIME\cite{ribeiro2016should}, faithfully explains model predictions by locally approximating them with an interpretable model. Szczepański et al.\cite{szczepanski2021new} applied LIME to BERT-based models fine-tuned for fake news detection, highlighting words and their weights to show their impact on prediction probability—higher weights increase the likelihood of a sentence being classified as fake, and vice versa. However, LIME is limited, failing to meet additive attribution properties like local accuracy, consistency, and missingness~\cite{lundberg2017shap}.

This issue is tackled by SHAP\cite{lundberg2017shap}, an effective feature attribution technique that uses Shapley values to measure unique additive feature importance. Kokalj et al.\cite{kokalj2021bert} extended SHAP for encoder-only models with TransSHAP, adapting it to BERT for tweet sentiment classification, visualizing predictions for positive and negative sentiments. The tool compares impact direction and magnitude per word, outperforming LIME in some visualization aspects, though LIME scored slightly higher in overall user preference. Challenges with SHAP include selecting feature removal methods and managing the computational complexity of estimating Shapley values~\cite{zhao2023explainability}.

\subsubsection{Classifier-based probing}
\label{classifier-based-probing-encoder-only}
Classifier-based probing trains lightweight classifiers on top of encoder-only models like BERT\cite{mohebbi2021exploring}, starting by freezing parameters and generating representations for input words, phrases, or sentences. Unlike feature attribution methods, which assess input feature contributions to predictions, this approach uses probing classifiers to identify linguistic properties or reasoning abilities\cite{zhao2023explainability}. It suits encoder-only models, pretrained with masked language modeling (MLM)~\cite{devlin2018bert}, producing robust contextualized embeddings for each token.


To explore word-level contextualized representations in encoder-only models like BERT, Tenney et al.\cite{tenney2019you} introduced edge probing, which examines how models encode sentence structure across syntactic, semantic, local, and long-range phenomena, targeting core NLP tasks like part-of-speech tagging\cite{tenney2019you}, semantic role labeling\cite{he-etal-2018-jointly}, and coreference resolution\cite{lee-etal-2018-higher}. Their study shows BERT outperforms non-contextual models, especially in syntactic tasks over semantic ones, indicating stronger encoding of syntactic features. However, edge probing lacks insight into layer-wise behavior and individual token functions~\cite{mohebbi2021exploring}.

To overcome this limitation, Mohebbi et al.\cite{mohebbi2021exploring} examine token representations in BERT’s space to explain performance trends across surface, syntactic, and semantic tasks, revealing that higher layers encode required knowledge, with most positional information diminishing through layers while sentence-ending tokens partly transmit it. They also find BERT encodes verb tense and noun number in these tokens, using a diagnostic classifier with attribution methods for qualitative insights. However, Hewitt and Liang\cite{hewitt2019designinginterpretingprobescontrol} question probing classifier faithfulness with control tasks, suggesting parameter-free probing—discussed next—as a solution.



\subsubsection{Parameter-free probing}

Parameter-free probing techniques bypass training external classifiers or adding parameters, directly analyzing outputs like activation patterns, attention weights, or contextualized embeddings from encoder-only models to detect linguistic structures\cite{wu2020perturbed}. This addresses a key flaw in classifier-based probing, where the probe may learn downstream task aspects, encoding them in its parameters rather than faithfully mirroring the model’s inherent knowledge\cite{hewitt2019designinginterpretingprobescontrol}.

Wu et al.~\cite{wu2020perturbed} introduce perturbed masking, a parameter-free probing technique to estimate inter-word correlations and extract global syntactic information. This method uses a two-stage process: first, masking a target word, then observing shifts in its contextualized representation when another word is masked, quantifying the second word’s influence. Impact matrices derived from BERT’s output capture these correlations, reflecting attention mechanisms’ dependency structures, though computed from outputs rather than intermediate representations. Algorithms extract syntactic trees from these matrices, showing BERT encodes rich syntactic properties, with the method also assessing its ability to model long document sequences.

Likewise, Zhou et al.\cite{zhou-srikumar-2021-directprobe} note that training classifiers as probes is unreliable due to varying representations requiring different classifiers, which can skew representation quality estimates. To counter this, they propose DIRECTPROBE\cite{zhou-srikumar-2021-directprobe}, a parameter-free heuristic method leveraging version space~\cite{mitchell1982generalization} to analyze embedding geometry, determining task suitability without classifiers.

\subsubsection{Attention-based methods}
\label{attention-based-methods-encoder-on}

Encoder-only models like BERT rely on transformer architectures with a fully attention-based, bidirectional self-attention mechanism\cite{devlin2018bert}. While prior XAI methods, such as feature attribution or lightweight classifier probing, assess input contributions, attention-based methods offer a direct view into model reasoning by visualizing self-attention distributions and token weight allocation. Various tools have been developed to leverage this, providing concise summaries and enhancing user interaction with LLMs\cite{hoover-etal-2020-exbert}.

To examine the attention mechanism, Vig et al.\cite{vig2019analyzing} developed BertViz, an open-source tool visualizing attention across multiple scales, offering perspectives via a high-level model view (showing all layers and heads) and a low-level neuron view (revealing neuron interactions in attention patterns). Applied to the encoder-only BERT model, it supports three use cases: detecting bias, identifying key attention heads, and linking neurons to behavior. However, relying solely on attention for faithful explanations can yield misleading results\cite{brunner2020identifiabilitytransformers}.

To address this, Hoover et al.\cite{hoover-etal-2020-exbert} introduced ExBERT, an interactive tool visualizing attention mechanisms and internal representations in transformer models. It features an attention view for examining aggregated patterns and a corpus view for summarizing hidden representations of selected tokens via statistics. Using attention visualization and nearest-neighbor search, ExBERT reveals captured information and potential biases in text inputs, though it is limited to local analyses, focusing on a few neighbors per token\cite{hoover-etal-2020-exbert}.

\subsection{XAI methods in decoder-only LLMs}

XAI methods for decoder-only LLMs address their autoregressive, left-to-right generation, presenting unique interpretability challenges and opportunities. Unlike encoder-only LLMs with stable, easily probed contextual embeddings, decoder-only models need techniques tailored to their unidirectional nature. Key methods include feature attribution-based approaches, in-context learning (ICL)-based methods, chain-of-thought (CoT) prompting, and mechanistic interpretability.

\subsubsection{Feature Attribution-based Methods}
\label{feature-attribution-decoder-only}

Feature attribution-based methods quantify individual input tokens' contributions to model outputs. As noted in Section\ref{feature-attribution-encoder-only}, they suit encoder-only models with bidirectional context. However, their application to decoder-only large language models (LLMs) is challenging due to the autoregressive, left-to-right token generation, where conventional bidirectional methods fail to capture the causal structure\cite{zhao2023explainability}. Recent efforts thus focus on developing attribution techniques tailored to this autoregressive decoding process.

Attributive Masking Learning (AML)\cite{barkan2024llm} uses a gradient-based approach to identify a minimal, sufficient input token subset preserving the model’s output distribution, unlike post-hoc approximations. AML employs a mask generator that learns binary token masks by minimizing Kullback-Leibler (KL) divergence\cite{kullback1951information} between full and masked predictions, offering faithful, causally grounded explanations via model gradients and internal activations. However, its dependence on internals limits use with closed-source LLMs like GPT-4\cite{openai2023gpt4} or Claude\cite{anthropic2025}, and its interpretability may falter in high-compositionality tasks (e.g., multi-hop reasoning) or ambiguous language due to unstable masking responses.

In contrast, Progressive Inference~\cite{kariyappa2024progressive} provides a parameter-free, model-agnostic method for black-box access, extending the input sequence token by token to track output probability evolution and build an influence curve for each prefix element. Aligned with autoregressive decoding, it explains generative behavior without needing model weights or gradients, broadening its use with proprietary models. However, its token-wise marginal attribution overlooks complex distant token dependencies, reducing explanatory power in tasks like summarization requiring global semantic understanding. Compared to AML, it sacrifices interpretive depth for accessibility, ideal for diagnostic cases with restricted model access.

\subsubsection{In-Context Learning (ICL)-based Methods}
\label{icl-based-methods}

In-context learning (ICL) enables decoder-only large language models (LLMs) like GPT-3 and GPT-4\cite{abramski2023cognitive} to perform novel tasks using a few prompt examples without parameter updates\cite{brown2020language}, a key feature of these architectures. However, its mechanisms are poorly understood due to the lack of gradient updates and opaque prompt processing. Thus, ICL explainability methods focus on interpreting how models extract and use prompt information, emphasizing structure, semantics, and token-level saliency.

To explore ICL behavior, Lampinen et al.~\cite{lampinen2022can} used contrastive prompts, substituting semantic content with syntactically coherent but meaningless tokens, revealing that models retain task performance without semantic cues. This suggests decoder-only LLMs depend more on syntactic structures and token positions than true semantic understanding, indicating that their generalization may arise from statistical pattern matching in prompt formats rather than abstract task comprehension.

Expanding on this, Li et al.~\cite{li2023saliencyicl} developed a saliency-guided method to examine how decoder-only LLMs interpret prompts at the token level, using output loss gradients to create saliency maps visualizing token influence. Their findings show attention focuses heavily on final tokens (e.g., answers or labels), suggesting models prioritize contrastive signals (e.g., correct vs. incorrect patterns) over holistic representations. This token-level analysis indicates ICL relies more on local alignment and positional regularities than deep semantic integration, underscoring the need to integrate prompt structure analysis with saliency-based explanations for a complete understanding.

\subsubsection{Chain-of-Thought (CoT) Prompting}
\label{cot-prompting}

Chain-of-thought (CoT) prompting adds intermediate reasoning steps to decoder-only large language models (LLMs) outputs, turning opaque predictions into interpretable rationales~\cite{wei2022chain, kojima2022large}. Aligned with autoregressive generation, CoT allows real-time tracking of reasoning paths, boosting performance on complex tasks like math or logic puzzles while embedding interpretability in outputs. Unlike in-context learning (ICL)-based methods, which depend on prompt structure for task behavior, CoT offers explicit reasoning insights in the response.

This makes CoT ideal for decoder-only models, whose left-to-right generation supports incremental, interpretable reasoning. A key method, SEA-CoT~\cite{zhang2024sea}, enhances logical consistency by using a natural language inference (NLI) model to verify each reasoning step against prior ones, filtering incoherent transitions for greater faithfulness and local coherence. However, its reliance on external NLI tools and focus on pairwise step relationships neglects global reasoning validity.

An alternative, Least-to-Most prompting~\cite{zhou2022least}, breaks complex problems into ordered subtasks (e.g., defining variables, setting equations, solving a math word problem) from easiest to hardest, fostering incremental, clear, and controllable reasoning chains. Unlike SEA-CoT’s post-hoc filtering, it structures reasoning during prompt creation. Though effective in logical and mathematical domains, it demands extensive prompt engineering and struggles with unstructured tasks like open-ended text generation. Together, these methods show how intermediate reasoning design—via validation or decomposition—enhances decoder-only model interpretability.

\subsubsection{Mechanistic Interpretability}
\label{mechanistic-interpretability}

Mechanistic interpretability methods explain decoder-only large language models (LLMs) by revealing internal computational structures driving behavior, tracing causal pathways to identify roles of specific neurons, attention heads, or feedforward layers in reasoning and prediction~\cite{olah2020zoom, nanda2023progress}. Ideal for autoregressive architectures with deeply stacked transformer blocks, this approach isolates modular components (e.g., subnetworks for specific functions). Unlike attribution-based or prompt-driven methods, it provides structural, fine-grained insights based on model parameters and activations.

A key technique, ROME (Rank-One Model Editing)\cite{meng2022locating}, offers a localized editing framework to modify factual knowledge in GPT-2\cite{radford2019language} by targeting specific feedforward layer neurons, showing some facts are stored in editable subnetworks (e.g., changing “Paris is the capital of France” to “Florida is the capital of France”). This links outputs to identifiable components, enhancing explainability and enabling direct edits. However, ROME is limited to factual recall, assumes fixed activation paths, and struggles with abstract or multi-hop reasoning requiring complex interactions.

Building on this, ACDC\cite{gurnee2024language} advances mechanistic interpretability by integrating activation patching, attribution, and pruning to identify minimal subgraphs driving broader cognitive tasks like analogical reasoning or number comparison. Related efforts, such as Path Patching\cite{wang2022interpretability} and Transformer Circuits~\cite{elhage2021mechanistic}, analyze attention and MLP contributions to trace signal propagation, uncovering reusable, modular components. However, these methods demand full white-box access and are computationally intensive, hindering scalability. Despite these limitations, they are vital for structurally grounded explanations and improving safe, controllable LLMs.

\sloppy
\subsection{XAI methods in encoder-decoder LLMs}
Encoder-decoder LLMs employ a two-part architecture with a cross-attention mechanism, integrating an encoder for understanding and a decoder for generation, enabling explainability at both stages. Unlike encoder-only models, they enhance generative capabilities, and compared to decoder-only models, they offer better input representation control. Key explainability methods include feature attribution, classifier-based methods, attention-based methods, and self-explanation techniques.

\sloppy
\subsubsection{Feature attribution-based methods}
\label{feature-attribution-encoder-decoder}

Feature attribution-based methods, as outlined in Sections\ref{feature-attribution-encoder-only} and\ref{feature-attribution-decoder-only}, assign relevance scores to each feature. Unlike their application in encoder-only and decoder-only LLMs, in encoder-decoder models, these methods evaluate token relevance across both input and output sequences\cite{enouen2024textgenshap}, reflecting the sequence-to-sequence nature by considering contributions from preceding output tokens as well as input tokens\cite{raffel2020exploring}.

To apply feature attribution-based methods for XAI in encoder-decoder models, Liu et al.\cite{liu2024reliabilityexplainabilitylanguagemodels} conducted a study on program generation, using a gradient-based SHAP\cite{lundberg2017shap} to interpret decision-making. They found models like CodeT5\cite{wang-etal-2023-codet5}, CodeReviewer\cite{li2022automatingcodereviewactivities}, and CodeT5+\cite{wang-etal-2023-codet5} prioritize keywords and identifiers over operators and separators, indicating strong code grammar and structure recognition. However, performance dropped significantly when removing even minor tokens, revealing robustness issues. Despite SHAP’s utility, its scalability struggles with large encoder-decoder models and long sequences, often causing impractical delays\cite{enouen2024textgenshap}.

To address this, Enouen et al.\cite{enouen2024textgenshap} introduced TextGenSHAP, a more efficient enhancement of Shapley values for large-scale encoder-decoder models like T5-large\cite{chung2022scalinginstructionfinetunedlanguagemodels}, T5-XXL\cite{chung2022scalinginstructionfinetunedlanguagemodels}, and T5-FiD\cite{izacard-grave-2021-leveraging}. Targeting open-ended text generation with long prompts, especially abstractive question answering with retrieval-augmented documents, TextGenSHAP improves recall in document retrieval systems. Though a major step forward in applying SHAP to large encoder-decoder models, computational costs for generating explanations remain a practical concern.

\subsubsection{Classifier-based methods}
As noted in Section\ref{classifier-based-probing-encoder-only}, classifier-based probing enhances explainability for encoder-only LLMs by training shallow classifiers on frozen representations to interpret predictions and behavior\cite{mohebbi2021exploring}. In encoder-decoder models, this approach uniquely targets both encoder and decoder components~\cite{raffel2020exploring}, addressing their complex interactions, which complicates interpretation.

To apply classifier-based probing to encoder-decoder LLMs, Koto et al.\cite{koto2021discourse} introduced document-level discourse probing to assess models like T5\cite{raffel2020exploring} and BART\cite{lewis2019bartdenoisingsequencetosequencepretraining} in capturing discourse relations, identifying active layers. The study reveals BART’s encoder layers excel in tasks like discourse connective\cite{nie-etal-2019-dissent}, RST nuclearity\cite{mann-thompson-1986-assertions}, and RST relation\cite{mann-thompson-1986-assertions}, while its decoder lags, focusing more on generation than understanding. Similar patterns in T5, a denoising autoencoder, show discourse knowledge is unevenly distributed across layers, shaped by pre-training objectives and architecture.

Extending classifier-based probing to implicit meaning, Li et al.~\cite{li2021implicit} explored how encoder-decoder LLMs like BART and T5 encode discourse-level information states, using a probing method to recover them. Results show implicit meanings are linearly decoded from entity mentions, capturing state changes and semantic details of final states, enhancing insight into these models' inner workings. However, limitations include recovering complete states only about half the time, with probing representations lacking the expressiveness for human-like language generation.


\subsubsection{Attention-based methods}

Attention-based methods leverage attention weights, especially cross-attention in encoder-decoder models, to interpret how input influences output generation\cite{raffel2020exploring}. As noted in Section\ref{attention-based-methods-encoder-on}, self-attention aids interpretability in encoder-only LLMs, but encoder-decoder models’ cross-attention layers directly connect encoder inputs to decoder outputs\cite{raffel2020exploring}. This makes attention a key tool for tasks like summarization or translation, where tracing source-to-target sequences is vital\cite{hoover-etal-2020-exbert, vig2019analyzing}, providing intuitive explanations by aligning attention distributions with semantic relevance.

To interpret encoder-decoder LLMs, Juraska et al.~\cite{juraska2021attention} introduced attention-guided decoding, analyzing cross-attention activations during decoding to identify source tokens influencing target tokens. By tracking attention weights across decoder steps and rescoring beams for semantic consistency, it reveals information flow from input to output, attributing output elements to specific inputs, aligning with human judgment in translation or summarization. However, its reliability depends on well-calibrated attention maps, as diffusion in lengthy or noisy inputs can diminish interpretive clarity, reducing explanation faithfulness.

In contrast, attention head masking~\cite{cao2021attention} uses an intervention-based approach, disabling specific attention heads at inference to assess their causal importance by observing output impacts, providing head-level attribution and insights into internal decision pathways rather than just token tracing. Though precise in isolating functional roles, it struggles with scalability due to combinatorial complexity and may miss distributed patterns where multiple weak heads collectively influence decisions.

\subsubsection{Self-explanation methods}

Self-explanation-based methods enhance encoder-decoder LLM interpretability by generating human-readable justifications with task predictions, unlike attention-based methods analyzing internal mechanisms\cite{juraska2021attention, cao2021attention}. Leveraging the decoder’s generative capacity\cite{narang2020wt5}, they produce explanations within the output sequence, aligning with encoder-decoder architectures as a unified text generation task\cite{rajani2019explain}. This offers an intuitive interface for assessing reasoning, especially in trust-demanding tasks like QA, classification, and commonsense inference\cite{narang2020wt5, yordanov2022few}, embedding reasoning into outputs for user-aligned interpretability without post-hoc analysis.

A key self-explanation method, WT5\cite{narang2020wt5}, builds on the T5 architecture, training models to output both answers and explanations, e.g., responding to “What is the capital of France?” with “Paris” and “The capital of France is Paris, so the answer is Paris.” This format links supporting facts to conclusions, boosting user trust and, in some cases, task performance via joint training\cite{narang2020wt5}. However, it relies on extensive explanation-annotated data, and it’s uncertain whether explanations reflect causal reasoning or just label patterns from training.

Extending this approach, Yordan et al.\cite{yordanov2022few} introduced Few-Shot Out-of-Domain NLE generation, targeting tasks with limited labeled explanations but abundant labels. Their method pretrains a model on a well-annotated source domain, then adapts it to a target domain with ample labels but scarce explanation data by aligning representation spaces and fine-tuning on limited examples, enabling quality explanation generation under data-scarce conditions\cite{yordanov2022few}. Unlike WT5’s rich supervision assumption, this enhances scalability across domains, though few-shot, knowledge-driven methods struggle with generalizing explanation quality and style, particularly in abstract reasoning.

\section{Challenges and Future Directions}
\label{sec:challenges-future}

Explainable artificial intelligence (XAI) for large language models (LLMs) is crucial for transparency and trust in high-stakes fields like healthcare and law\cite{healthxai,hutchinson2020accountability}. Despite progress in XAI methods for encoder-only, decoder-only, and encoder-decoder architectures (Section\ref{sec:XAI_methods}), notable challenges remain, requiring innovative solutions. Below, we outline four key challenges and future directions to enhance explainable LLMs.

\textbf{(1) Explanation and Emergent Abilities}
Ground truth explanations for LLMs are typically inaccessible, with no benchmark datasets to evaluate global explanations of individual components. This hinders assessing explanation faithfulness and fidelity, especially for emergent abilities that arise as models scale and training data grows, even without explicit training for such tasks. Open questions include: (1) How can we design algorithms to reflect LLM decision-making without ground truth? (2) What model architectures and data characteristics drive emergent skills in LLMs like ChatGPT or LLaMA? Solutions may involve human evaluations, synthetic datasets, and investigating the interplay between model design and training data~\cite{luo2024understandingutilizationsurveyexplainability}.

\textbf{(2) Paradigm Comparison and Shortcut Learning}
The fine-tuning and prompting paradigms show divergent performance in natural language inference (NLI) and out-of-distribution (OOD) tasks, relying on distinct reasoning strategies. Recent studies reveal LLMs use shortcuts like lexical or position bias in fine-tuning and falter with long contexts in prompting. Key questions are: (1) How do their prediction rationales differ? (2) Can OOD robustness be linked to reasoning variations? Tackling this involves comparing explanatory differences and reducing shortcut dependence to enhance generalization~\cite{luo2024understandingutilizationsurveyexplainability}.

\textbf{(3) Interpretability Efficiency and Redundancy}
High-dimensional representations and deep transformer layers in LLMs like BERT\cite{devlin2018bert} and GPT-4\cite{openai2023gpt4} render decision tracing computationally intensive. XAI methods like SHAP\cite{lundberg2017shap,enouen2024textgenshap} require substantial resources, hindering scalability. Additionally, attention redundancy in models like BERT-base and OPT-66B indicates pruning redundant heads could boost efficiency with minimal performance impact. Future efforts should prioritize efficient XAI techniques and model compression strategies\cite{luo2024understandingutilizationsurveyexplainability}.

\textbf{(4) Ethics and Temporal Analysis}
The lack of interpretability raises ethical risks like bias and misinformation, requiring audits and alignment with human values. Tools like BertViz\cite{vig2019analyzing} expose biases, but research overlooks training dynamics, depending on post-hoc analysis. Temporal studies, such as detecting Syntactic Attention Structure (SAS) in pre-training, indicate dynamic analysis could reveal causal patterns. Emphasizing interpretability with scale and safety is vital for ethical LLM development\cite{luo2024understandingutilizationsurveyexplainability}.

\section{Conclusion}
\label{sec:conclusion}

Large language models have transformed natural language processing, but their opaque nature hinders transparency and trust in critical domains like healthcare and law. This survey offers a comprehensive review of explainable AI methods for these models, introducing a novel taxonomy based on encoder-only, decoder-only, and encoder-decoder architectures, and examining techniques such as feature attribution, probing, attention-based methods, and chain-of-thought prompting. Despite progress, challenges in complexity, standardization, generalization, efficiency, and ethics underscore the need for innovative solutions to achieve scalable, faithful, and ethical explanations, guiding future efforts toward transparent, responsible, and trustworthy models.

\bibliographystyle{ACM-Reference-Format}
\bibliography{sample-base2}


\begin{thebibliography}{71}


\ifx \showCODEN    \undefined \def \showCODEN     #1{\unskip}     \fi
\ifx \showDOI      \undefined \def \showDOI       #1{#1}\fi
\ifx \showISBNx    \undefined \def \showISBNx     #1{\unskip}     \fi
\ifx \showISBNxiii \undefined \def \showISBNxiii  #1{\unskip}     \fi
\ifx \showISSN     \undefined \def \showISSN      #1{\unskip}     \fi
\ifx \showLCCN     \undefined \def \showLCCN      #1{\unskip}     \fi
\ifx \shownote     \undefined \def \shownote      #1{#1}          \fi
\ifx \showarticletitle \undefined \def \showarticletitle #1{#1}   \fi
\ifx \showURL      \undefined \def \showURL       {\relax}        \fi
\providecommand\bibfield[2]{#2}
\providecommand\bibinfo[2]{#2}
\providecommand\natexlab[1]{#1}
\providecommand\showeprint[2][]{arXiv:#2}

\bibitem[A. and et~al.(2023a)]%
        {healthxai}
\bibfield{author}{\bibinfo{person}{Chaddad A.} {and} \bibinfo{person}{et al.}} \bibinfo{year}{2023}\natexlab{a}.
\newblock \showarticletitle{Survey of Explainable AI Techniques in Healthcare}.
\newblock \bibinfo{journal}{\emph{Sensors}}  \bibinfo{volume}{23} (\bibinfo{date}{01} \bibinfo{year}{2023}), \bibinfo{pages}{634}.
\newblock


\bibitem[A. and et~al.(2023b)]%
        {abramski2023cognitive}
\bibfield{author}{\bibinfo{person}{Katherine A.} {and} \bibinfo{person}{et al.}} \bibinfo{year}{2023}\natexlab{b}.
\newblock \showarticletitle{Cognitive network science reveals bias in gpt-3, gpt-3.5 turbo, and gpt-4 mirroring math anxiety in high-school students}.
\newblock \bibinfo{journal}{\emph{Big Data and Cognitive Computing}} \bibinfo{volume}{7}, \bibinfo{number}{3} (\bibinfo{year}{2023}), \bibinfo{pages}{124}.
\newblock


\bibitem[A. and et~al.(2020)]%
        {lopes2020litetrainingstrategiesportugueseenglish}
\bibfield{author}{\bibinfo{person}{Lopes A.} {and} \bibinfo{person}{et al.}} \bibinfo{year}{2020}\natexlab{}.
\newblock \bibinfo{title}{Lite Training Strategies for Portuguese-English and English-Portuguese Translation}.
\newblock
\newblock
\showeprint{2008.08769 [cs.CL]}


\bibitem[A. and et~al.(2019)]%
        {nie-etal-2019-dissent}
\bibfield{author}{\bibinfo{person}{Nie A.} {and} \bibinfo{person}{et al.}} \bibinfo{year}{2019}\natexlab{}.
\newblock \showarticletitle{DisSent: Learning Sentence Representations from Explicit Discourse Relations}. In \bibinfo{booktitle}{\emph{ACL}}. \bibinfo{pages}{4497--4510}.
\newblock


\bibitem[A. and et~al.(2017)]%
        {vaswani2017attention}
\bibfield{author}{\bibinfo{person}{Vaswani A.} {and} \bibinfo{person}{et al.}} \bibinfo{year}{2017}\natexlab{}.
\newblock \showarticletitle{Attention is all you need}. In \bibinfo{booktitle}{\emph{NeurIPS}}. \bibinfo{pages}{30}.
\newblock


\bibitem[Anthropic(2025)]%
        {anthropic2025}
\bibfield{author}{\bibinfo{person}{Anthropic}.} \bibinfo{year}{2025}\natexlab{}.
\newblock \bibinfo{title}{Official Documentation for Anthropic APIs and Models}.
\newblock
\newblock
\newblock
\shownote{Accessed on June 16, 2025}.


\bibitem[B. and et~al(2020)]%
        {hutchinson2020accountability}
\bibfield{author}{\bibinfo{person}{Hutchinson B.} {and} \bibinfo{person}{et al}.} \bibinfo{year}{2020}\natexlab{}.
\newblock \bibinfo{title}{Accountability of AI under the law: The role of explanation}.
\newblock
\newblock
\showeprint{1711.01134 [cs.AI]}


\bibitem[B. and et~al.(2020a)]%
        {hoover-etal-2020-exbert}
\bibfield{author}{\bibinfo{person}{Hoover B.} {and} \bibinfo{person}{et al.}} \bibinfo{year}{2020}\natexlab{a}.
\newblock \showarticletitle{exBERT: A Visual Analysis Tool to Explore Learned Representations in Transformer Models}. In \bibinfo{booktitle}{\emph{ACL}}. \bibinfo{pages}{187--196}.
\newblock


\bibitem[B. and et~al.(2024)]%
        {barkan2024llm}
\bibfield{author}{\bibinfo{person}{Oren B.} {and} \bibinfo{person}{et al.}} \bibinfo{year}{2024}\natexlab{}.
\newblock \showarticletitle{LLM Explainability via Attributive Masking Learning}. In \bibinfo{booktitle}{\emph{EMNLP}}. \bibinfo{pages}{9522--9537}.
\newblock


\bibitem[B. and et~al.(2020b)]%
        {brown2020language}
\bibfield{author}{\bibinfo{person}{Tom B.} {and} \bibinfo{person}{et al.}} \bibinfo{year}{2020}\natexlab{b}.
\newblock \showarticletitle{Language models are few-shot learners}.
\newblock \bibinfo{journal}{\emph{NeurIPS}}  \bibinfo{volume}{33} (\bibinfo{year}{2020}), \bibinfo{pages}{1877--1901}.
\newblock


\bibitem[C. and et~al(2024)]%
        {ng2024educational}
\bibfield{author}{\bibinfo{person}{Ng C.} {and} \bibinfo{person}{et al}.} \bibinfo{year}{2024}\natexlab{}.
\newblock \bibinfo{title}{Educational personalized learning path planning with large language models}.
\newblock
\newblock
\showeprint{2407.11773 [cs.AI]}


\bibitem[C. and et~al.(2019)]%
        {rudin2019stop}
\bibfield{author}{\bibinfo{person}{Rudin C.} {and} \bibinfo{person}{et al.}} \bibinfo{year}{2019}\natexlab{}.
\newblock \showarticletitle{Stop explaining black box machine learning models for high stakes decisions and use interpretable models instead}.
\newblock \bibinfo{journal}{\emph{Nature Machine Intelligence}} \bibinfo{volume}{1}, \bibinfo{number}{5} (\bibinfo{year}{2019}), \bibinfo{pages}{206--215}.
\newblock


\bibitem[C. and et~al.(2023)]%
        {raffel2020exploring}
\bibfield{author}{\bibinfo{person}{Raffel C.} {and} \bibinfo{person}{et al.}} \bibinfo{year}{2023}\natexlab{}.
\newblock \bibinfo{title}{Exploring the Limits of Transfer Learning with a Unified Text-to-Text Transformer}.
\newblock
\newblock
\showeprint{1910.10683 [cs.LG]}


\bibitem[D. and et~al.(2022)]%
        {zhou2022least}
\bibfield{author}{\bibinfo{person}{Zhou D.} {and} \bibinfo{person}{et al.}} \bibinfo{year}{2022}\natexlab{}.
\newblock \showarticletitle{Least-to-most prompting enables complex reasoning in large language models}.
\newblock \bibinfo{journal}{\emph{ICLR}} (\bibinfo{year}{2022}).
\newblock


\bibitem[D. and et~al.(2024)]%
        {zhang2024sea}
\bibfield{author}{\bibinfo{person}{Zhang D.} {and} \bibinfo{person}{et al.}} \bibinfo{year}{2024}\natexlab{}.
\newblock \showarticletitle{SEA-CoT: Self-Entailment-Alignment Chain-of-Thought Prompting for Explainable Reasoning}. In \bibinfo{booktitle}{\emph{NAACL}}.
\newblock


\bibitem[E. and et~al.(2021)]%
        {kokalj2021bert}
\bibfield{author}{\bibinfo{person}{Kokalj E.} {and} \bibinfo{person}{et al.}} \bibinfo{year}{2021}\natexlab{}.
\newblock \showarticletitle{BERT meets shapley: Extending SHAP explanations to transformer-based classifiers}. In \bibinfo{booktitle}{\emph{EACL Hackashop}}. \bibinfo{pages}{16--21}.
\newblock


\bibitem[F. and et~al.(2021)]%
        {koto2021discourse}
\bibfield{author}{\bibinfo{person}{Koto F.} {and} \bibinfo{person}{et al.}} \bibinfo{year}{2021}\natexlab{}.
\newblock \showarticletitle{Discourse Probing of Pretrained Language Models}. In \bibinfo{booktitle}{\emph{NAACL-HLT}}. \bibinfo{pages}{3849--3864}.
\newblock


\bibitem[F. and et~al.(2019)]%
        {rajani2019explain}
\bibfield{author}{\bibinfo{person}{Rajani~N. F.} {and} \bibinfo{person}{et al.}} \bibinfo{year}{2019}\natexlab{}.
\newblock \showarticletitle{Explain Yourself! Leveraging Language Models for Commonsense Reasoning}. In \bibinfo{booktitle}{\emph{ACL}}. \bibinfo{pages}{4932--4942}.
\newblock


\bibitem[G. and et~al.(2020)]%
        {brunner2020identifiabilitytransformers}
\bibfield{author}{\bibinfo{person}{Brunner G.} {and} \bibinfo{person}{et al.}} \bibinfo{year}{2020}\natexlab{}.
\newblock \bibinfo{title}{On Identifiability in Transformers}.
\newblock
\newblock
\showeprint{1908.04211 [cs.CL]}


\bibitem[G. and et~al.(2021)]%
        {izacard-grave-2021-leveraging}
\bibfield{author}{\bibinfo{person}{Izacard G.} {and} \bibinfo{person}{et al.}} \bibinfo{year}{2021}\natexlab{}.
\newblock \showarticletitle{Leveraging Passage Retrieval with Generative Models for Open Domain Question Answering}. In \bibinfo{booktitle}{\emph{EACL}}. \bibinfo{pages}{874--880}.
\newblock


\bibitem[H. and et~al.(2022)]%
        {chung2022scalinginstructionfinetunedlanguagemodels}
\bibfield{author}{\bibinfo{person}{Chung H.} {and} \bibinfo{person}{et al.}} \bibinfo{year}{2022}\natexlab{}.
\newblock \bibinfo{title}{Scaling Instruction-Finetuned Language Models}.
\newblock
\newblock
\showeprint{2210.11416}


\bibitem[H. and et~al.(2024a)]%
        {luo2024understandingutilizationsurveyexplainability}
\bibfield{author}{\bibinfo{person}{Luo H.} {and} \bibinfo{person}{et al.}} \bibinfo{year}{2024}\natexlab{a}.
\newblock \bibinfo{title}{From Understanding to Utilization: A Survey on Explainability for Large Language Models}.
\newblock
\newblock
\showeprint{2401.12874 [cs.CL]}


\bibitem[H. and et~al.(2021)]%
        {mohebbi2021exploring}
\bibfield{author}{\bibinfo{person}{Mohebbi H.} {and} \bibinfo{person}{et al.}} \bibinfo{year}{2021}\natexlab{}.
\newblock \showarticletitle{Exploring the Role of BERT Token Representations to Explain Sentence Probing Results}. In \bibinfo{booktitle}{\emph{EMNLP}}. \bibinfo{pages}{792--806}.
\newblock


\bibitem[H. and et~al(2023)]%
        {touvron2023llama}
\bibfield{author}{\bibinfo{person}{Touvron H.} {and} \bibinfo{person}{et al}.} \bibinfo{year}{2023}\natexlab{}.
\newblock \bibinfo{title}{Llama 2: Open foundation and fine-tuned chat models}.
\newblock
\newblock
\showeprint{2307.09288 [cs.AI]}


\bibitem[H. and et~al.(2024b)]%
        {zhao2023explainability}
\bibfield{author}{\bibinfo{person}{Zhao H.} {and} \bibinfo{person}{et al.}} \bibinfo{year}{2024}\natexlab{b}.
\newblock \showarticletitle{Explainability for large language models: A survey}.
\newblock \bibinfo{journal}{\emph{ACM Transactions on Intelligent Systems and Technology}} \bibinfo{volume}{15}, \bibinfo{number}{2} (\bibinfo{year}{2024}), \bibinfo{pages}{1--38}.
\newblock


\bibitem[I. and et~al.(2019)]%
        {tenney2019you}
\bibfield{author}{\bibinfo{person}{Tenney I.} {and} \bibinfo{person}{et al.}} \bibinfo{year}{2019}\natexlab{}.
\newblock \showarticletitle{What do you learn from context? Probing for sentence structure in contextualized word representations}. In \bibinfo{booktitle}{\emph{ICLR}}.
\newblock


\bibitem[J. and et~al(2023)]%
        {openai2023gpt4}
\bibfield{author}{\bibinfo{person}{Achiam J.} {and} \bibinfo{person}{et al}.} \bibinfo{year}{2023}\natexlab{}.
\newblock \bibinfo{title}{Gpt-4 technical report}.
\newblock
\newblock
\showeprint{2303.08774}


\bibitem[J. and et~al(2018)]%
        {devlin2018bert}
\bibfield{author}{\bibinfo{person}{Devlin J.} {and} \bibinfo{person}{et al}.} \bibinfo{year}{2018}\natexlab{}.
\newblock \bibinfo{title}{BERT: Pre-training of deep bidirectional transformers for language understanding}.
\newblock
\newblock
\showeprint{1810.04805 [cs.CL]}


\bibitem[J. and et~al.(2024)]%
        {enouen2024textgenshap}
\bibfield{author}{\bibinfo{person}{Enouen J.} {and} \bibinfo{person}{et al.}} \bibinfo{year}{2024}\natexlab{}.
\newblock \showarticletitle{TextGenSHAP: Scalable Post-Hoc Explanations in Text Generation with Long Documents}. In \bibinfo{booktitle}{\emph{ACL}}. \bibinfo{pages}{13984--14011}.
\newblock


\bibitem[J. and et~al.(2019a)]%
        {hewitt2019designinginterpretingprobescontrol}
\bibfield{author}{\bibinfo{person}{Hewitt J.} {and} \bibinfo{person}{et al.}} \bibinfo{year}{2019}\natexlab{a}.
\newblock \bibinfo{title}{Designing and Interpreting Probes with Control Tasks}.
\newblock
\newblock
\showeprint{1909.03368 [cs.CL]}


\bibitem[J. and et~al.(2021)]%
        {juraska2021attention}
\bibfield{author}{\bibinfo{person}{Juraj J.} {and} \bibinfo{person}{et al.}} \bibinfo{year}{2021}\natexlab{}.
\newblock \showarticletitle{Attention Is Indeed All You Need: Semantically Attention-Guided Decoding for Data-to-Text NLG}. In \bibinfo{booktitle}{\emph{INLG}}. \bibinfo{pages}{416--431}.
\newblock


\bibitem[J. and et~al.(2020)]%
        {lee2020biobert}
\bibfield{author}{\bibinfo{person}{Lee J.} {and} \bibinfo{person}{et al.}} \bibinfo{year}{2020}\natexlab{}.
\newblock \showarticletitle{BioBERT: A Pre-trained Biomedical Language Representation Model for Biomedical Text Mining}.
\newblock \bibinfo{journal}{\emph{Bioinformatics}} \bibinfo{volume}{36}, \bibinfo{number}{4} (\bibinfo{year}{2020}), \bibinfo{pages}{1234--1240}.
\newblock


\bibitem[J. and et~al.(2019b)]%
        {vig2019analyzing}
\bibfield{author}{\bibinfo{person}{Vig J.} {and} \bibinfo{person}{et al.}} \bibinfo{year}{2019}\natexlab{b}.
\newblock \showarticletitle{A Multiscale Visualization of Attention in the Transformer Model}. In \bibinfo{booktitle}{\emph{ACL}}. \bibinfo{pages}{37--42}.
\newblock


\bibitem[J. and et~al.(2022)]%
        {wei2022chain}
\bibfield{author}{\bibinfo{person}{Wei J.} {and} \bibinfo{person}{et al.}} \bibinfo{year}{2022}\natexlab{}.
\newblock \showarticletitle{Chain-of-thought prompting elicits reasoning in large language models}.
\newblock \bibinfo{journal}{\emph{NeurIPS}}  \bibinfo{volume}{35} (\bibinfo{year}{2022}), \bibinfo{pages}{24824--24837}.
\newblock


\bibitem[J. and et~al(2022)]%
        {wei2022emergent}
\bibfield{author}{\bibinfo{person}{Wei J.} {and} \bibinfo{person}{et al}.} \bibinfo{year}{2022}\natexlab{}.
\newblock \bibinfo{title}{Emergent abilities of large language models}.
\newblock
\newblock
\showeprint{2206.07682 [cs.AI]}


\bibitem[J. and et~al.(2022)]%
        {yu2022interaction}
\bibfield{author}{\bibinfo{person}{Yu J.} {and} \bibinfo{person}{et al.}} \bibinfo{year}{2022}\natexlab{}.
\newblock \bibinfo{title}{INTERACTION: A Generative XAI Framework for Natural Language Inference Explanations}.
\newblock
\newblock
\showeprint{2209.01061 [cs.CL]}


\bibitem[K. and et~al.(2018)]%
        {lee-etal-2018-higher}
\bibfield{author}{\bibinfo{person}{Lee K.} {and} \bibinfo{person}{et al.}} \bibinfo{year}{2018}\natexlab{}.
\newblock \showarticletitle{Higher-Order Coreference Resolution with Coarse-to-Fine Inference}. In \bibinfo{booktitle}{\emph{NAACL-HLT}}. \bibinfo{pages}{687--692}.
\newblock


\bibitem[K. and et~al(2022)]%
        {lampinen2022can}
\bibfield{author}{\bibinfo{person}{Lampinen K.} {and} \bibinfo{person}{et al}.} \bibinfo{year}{2022}\natexlab{}.
\newblock \bibinfo{title}{Can language models learn from explanations in context?}
\newblock
\newblock
\showeprint{2204.02329 [cs.AI]}


\bibitem[K. and et~al.(2022)]%
        {meng2022locating}
\bibfield{author}{\bibinfo{person}{Meng K.} {and} \bibinfo{person}{et al.}} \bibinfo{year}{2022}\natexlab{}.
\newblock \showarticletitle{Locating and editing factual associations in GPT}. In \bibinfo{booktitle}{\emph{NeurIPS}}. \bibinfo{pages}{17359--17372}.
\newblock


\bibitem[K. and et~al.(1951)]%
        {kullback1951information}
\bibfield{author}{\bibinfo{person}{Solomon K.} {and} \bibinfo{person}{et al.}} \bibinfo{year}{1951}\natexlab{}.
\newblock \showarticletitle{On Information and Sufficiency}.
\newblock \bibinfo{journal}{\emph{The Annals of Mathematical Statistics}} \bibinfo{volume}{22}, \bibinfo{number}{1} (\bibinfo{date}{March} \bibinfo{year}{1951}), \bibinfo{pages}{79--86}.
\newblock


\bibitem[L. and et~al.(2023)]%
        {lampinen2023complementary}
\bibfield{author}{\bibinfo{person}{Andrew L.} {and} \bibinfo{person}{et al.}} \bibinfo{year}{2023}\natexlab{}.
\newblock \showarticletitle{Complementary Explanations: In-Context Learning via Composition of Natural Language and Computation Traces}. In \bibinfo{booktitle}{\emph{ACL}}.
\newblock


\bibitem[L. and et~al.(2018)]%
        {he-etal-2018-jointly}
\bibfield{author}{\bibinfo{person}{He L.} {and} \bibinfo{person}{et al.}} \bibinfo{year}{2018}\natexlab{}.
\newblock \showarticletitle{Jointly Predicting Predicates and Arguments in Neural Semantic Role Labeling}. In \bibinfo{booktitle}{\emph{ACL}}. \bibinfo{pages}{364--369}.
\newblock


\bibitem[M. and et~al(2021)]%
        {chen2021evaluating}
\bibfield{author}{\bibinfo{person}{Chen M.} {and} \bibinfo{person}{et al}.} \bibinfo{year}{2021}\natexlab{}.
\newblock \bibinfo{title}{Evaluating large language models trained on code}.
\newblock
\newblock
\showeprint{2107.03374 [cs.LG]}


\bibitem[M. and et~al.(2019)]%
        {lewis2019bartdenoisingsequencetosequencepretraining}
\bibfield{author}{\bibinfo{person}{Lewis M.} {and} \bibinfo{person}{et al.}} \bibinfo{year}{2019}\natexlab{}.
\newblock \bibinfo{title}{BART: Denoising Sequence-to-Sequence Pre-training for Natural Language Generation, Translation, and Comprehension}.
\newblock
\newblock
\showeprint{1910.13461 [cs.CL]}


\bibitem[M. and et~al.(2017)]%
        {lundberg2017shap}
\bibfield{author}{\bibinfo{person}{Lundberg~S. M.} {and} \bibinfo{person}{et al.}} \bibinfo{year}{2017}\natexlab{}.
\newblock \showarticletitle{A Unified Approach to Interpreting Model Predictions}. In \bibinfo{booktitle}{\emph{NeurIPS}}.
\newblock


\bibitem[M. and et~al.(1982)]%
        {mitchell1982generalization}
\bibfield{author}{\bibinfo{person}{Mitchell~T. M.} {and} \bibinfo{person}{et al.}} \bibinfo{year}{1982}\natexlab{}.
\newblock \showarticletitle{Generalization as search}.
\newblock \bibinfo{journal}{\emph{Artificial Intelligence}} \bibinfo{volume}{18}, \bibinfo{number}{2} (\bibinfo{year}{1982}), \bibinfo{pages}{203--226}.
\newblock


\bibitem[M. and et~al.(2021)]%
        {szczepanski2021new}
\bibfield{author}{\bibinfo{person}{Szczepański M.} {and} \bibinfo{person}{et al.}} \bibinfo{year}{2021}\natexlab{}.
\newblock \showarticletitle{New explainability method for BERT-based model in fake news detection}.
\newblock \bibinfo{journal}{\emph{Scientific Reports}} \bibinfo{volume}{11}, \bibinfo{number}{1} (\bibinfo{year}{2021}), \bibinfo{pages}{23705}.
\newblock


\bibitem[N. and et~al.(2023)]%
        {nanda2023progress}
\bibfield{author}{\bibinfo{person}{Nanda N.} {and} \bibinfo{person}{et al.}} \bibinfo{year}{2023}\natexlab{}.
\newblock \showarticletitle{Progress measures for grokking via mechanistic interpretability}. In \bibinfo{booktitle}{\emph{ICLR}}.
\newblock


\bibitem[O. and et~al.(2020)]%
        {olah2020zoom}
\bibfield{author}{\bibinfo{person}{Chris O.} {and} \bibinfo{person}{et al.}} \bibinfo{year}{2020}\natexlab{}.
\newblock \showarticletitle{Zoom in: An introduction to circuits}.
\newblock \bibinfo{journal}{\emph{Distill}} \bibinfo{volume}{5}, \bibinfo{number}{3} (\bibinfo{year}{2020}), \bibinfo{pages}{e00024--001}.
\newblock


\bibitem[O. and et~al.(2022)]%
        {elhage2021mechanistic}
\bibfield{author}{\bibinfo{person}{Catherine O.} {and} \bibinfo{person}{et al.}} \bibinfo{year}{2022}\natexlab{}.
\newblock \bibinfo{title}{In-context learning and induction heads}.
\newblock
\newblock
\showeprint{2209.11895 [cs.CL]}


\bibitem[P. and et~al.(2024)]%
        {mavrepis2024xai}
\bibfield{author}{\bibinfo{person}{Mavrepis P.} {and} \bibinfo{person}{et al.}} \bibinfo{year}{2024}\natexlab{}.
\newblock \bibinfo{title}{XAI for All: Can Large Language Models Simplify Explainable AI?}
\newblock
\newblock
\showeprint{2401.13110 [cs.AI]}


\bibitem[R. and et~al.(2019)]%
        {radford2019language}
\bibfield{author}{\bibinfo{person}{Alec R.} {and} \bibinfo{person}{et al.}} \bibinfo{year}{2019}\natexlab{}.
\newblock \showarticletitle{Language models are unsupervised multitask learners}.
\newblock \bibinfo{journal}{\emph{OpenAI blog}} \bibinfo{volume}{1}, \bibinfo{number}{8} (\bibinfo{year}{2019}), \bibinfo{pages}{9}.
\newblock


\bibitem[R. and et~al(2021)]%
        {bommasani2021opportunities}
\bibfield{author}{\bibinfo{person}{Bommasani R.} {and} \bibinfo{person}{et al}.} \bibinfo{year}{2021}\natexlab{}.
\newblock \bibinfo{title}{On the opportunities and risks of foundation models}.
\newblock
\newblock
\showeprint{2108.07258 [cs.AI]}


\bibitem[R. and et~al.(2022)]%
        {wang2022interpretability}
\bibfield{author}{\bibinfo{person}{Wang R.} {and} \bibinfo{person}{et al.}} \bibinfo{year}{2022}\natexlab{}.
\newblock \showarticletitle{Interpretability in the Wild: a Circuit for Indirect Object Identification in GPT-2 Small}. In \bibinfo{booktitle}{\emph{ICLR}}.
\newblock


\bibitem[S. and et~al.(2021)]%
        {cao2021attention}
\bibfield{author}{\bibinfo{person}{Cao S.} {and} \bibinfo{person}{et al.}} \bibinfo{year}{2021}\natexlab{}.
\newblock \showarticletitle{Attention Head Masking for Inference Time Content Selection in Abstractive Summarization}. In \bibinfo{booktitle}{\emph{NAACL-HLT}}. \bibinfo{pages}{5008--5016}.
\newblock


\bibitem[S. and et~al.(2024)]%
        {kariyappa2024progressive}
\bibfield{author}{\bibinfo{person}{Kariyappa S.} {and} \bibinfo{person}{et al.}} \bibinfo{year}{2024}\natexlab{}.
\newblock \showarticletitle{Progressive Inference: Explaining Decoder-Only Sequence Classification Models Using Intermediate Predictions}. In \bibinfo{booktitle}{\emph{ICML}}. \bibinfo{pages}{23238--23255}.
\newblock


\bibitem[S. and et~al.(2023)]%
        {li2023saliencyicl}
\bibfield{author}{\bibinfo{person}{Li S.} {and} \bibinfo{person}{et al.}} \bibinfo{year}{2023}\natexlab{}.
\newblock \showarticletitle{Analyzing In-Context Learning with Saliency Methods: A Case Study on Contrastive Demonstrations}.
\newblock  (\bibinfo{year}{2023}).
\newblock
\showeprint{2307.05052 [cs.CL]}


\bibitem[S. and et~al.(2020)]%
        {narang2020wt5}
\bibfield{author}{\bibinfo{person}{Narang S.} {and} \bibinfo{person}{et al.}} \bibinfo{year}{2020}\natexlab{}.
\newblock \bibinfo{title}{Wt5?! training text-to-text models to explain their predictions}.
\newblock
\newblock
\showeprint{2004.14546 [cs.CL]}


\bibitem[T. and et~al.(2022)]%
        {kojima2022large}
\bibfield{author}{\bibinfo{person}{Kojima T.} {and} \bibinfo{person}{et al.}} \bibinfo{year}{2022}\natexlab{}.
\newblock \showarticletitle{Large language models are zero-shot reasoners}.
\newblock \bibinfo{journal}{\emph{NeurIPS}}  \bibinfo{volume}{35} (\bibinfo{year}{2022}), \bibinfo{pages}{22199--22213}.
\newblock


\bibitem[T. and et~al.(2016)]%
        {ribeiro2016should}
\bibfield{author}{\bibinfo{person}{Ribeiro~M. T.} {and} \bibinfo{person}{et al.}} \bibinfo{year}{2016}\natexlab{}.
\newblock \showarticletitle{"Why should I trust you?": Explaining the predictions of any classifier}. In \bibinfo{booktitle}{\emph{KDD}}. \bibinfo{pages}{1135--1144}.
\newblock


\bibitem[W. and et~al.(2024)]%
        {gurnee2024language}
\bibfield{author}{\bibinfo{person}{Gurnee W.} {and} \bibinfo{person}{et al.}} \bibinfo{year}{2024}\natexlab{}.
\newblock \showarticletitle{Language Models Represent Space and Time}.
\newblock \bibinfo{journal}{\emph{ICLR}}.
\newblock


\bibitem[W. and et~al.(1986)]%
        {mann-thompson-1986-assertions}
\bibfield{author}{\bibinfo{person}{Mann W.} {and} \bibinfo{person}{et al.}} \bibinfo{year}{1986}\natexlab{}.
\newblock \showarticletitle{Assertions from Discourse Structure}. In \bibinfo{booktitle}{\emph{SC-NL Workshop}}.
\newblock


\bibitem[Y. and et~al.(2024a)]%
        {dang2025multimodal}
\bibfield{author}{\bibinfo{person}{Dang Y.} {and} \bibinfo{person}{et al.}} \bibinfo{year}{2024}\natexlab{a}.
\newblock \bibinfo{title}{Explainable and Interpretable Multimodal Large Language Models: A Comprehensive Survey}.
\newblock
\newblock
\showeprint{2412.02104}


\bibitem[Y. and et~al.(2024b)]%
        {liu2024reliabilityexplainabilitylanguagemodels}
\bibfield{author}{\bibinfo{person}{Liu Y.} {and} \bibinfo{person}{et al.}} \bibinfo{year}{2024}\natexlab{b}.
\newblock \showarticletitle{On the Reliability and Explainability of Language Models for Program Generation}.
\newblock \bibinfo{journal}{\emph{ACM Trans. Softw. Eng. Methodol.}} \bibinfo{volume}{33}, \bibinfo{number}{5}, Article \bibinfo{articleno}{126} (\bibinfo{year}{2024}), \bibinfo{numpages}{26}~pages.
\newblock
\showISSN{1049-331X}


\bibitem[Y. and et~al.(2023)]%
        {wang-etal-2023-codet5}
\bibfield{author}{\bibinfo{person}{Wang Y.} {and} \bibinfo{person}{et al.}} \bibinfo{year}{2023}\natexlab{}.
\newblock \showarticletitle{CodeT5+: Open Code Large Language Models for Code Understanding and Generation}. In \bibinfo{booktitle}{\emph{EMNLP}}. \bibinfo{pages}{1069--1088}.
\newblock


\bibitem[Y. and et~al.(2022)]%
        {yordanov2022few}
\bibfield{author}{\bibinfo{person}{Yordan Y.} {and} \bibinfo{person}{et al.}} \bibinfo{year}{2022}\natexlab{}.
\newblock \showarticletitle{Few-Shot Out-of-Domain Transfer Learning of Natural Language Explanations in a Label-Abundant Setup}. In \bibinfo{booktitle}{\emph{EMNLP}}. \bibinfo{pages}{3486--3501}.
\newblock


\bibitem[Y. and et~al.(2021)]%
        {zhou-srikumar-2021-directprobe}
\bibfield{author}{\bibinfo{person}{Zhou Y.} {and} \bibinfo{person}{et al.}} \bibinfo{year}{2021}\natexlab{}.
\newblock \showarticletitle{DirectProbe: Studying Representations without Classifiers}. In \bibinfo{booktitle}{\emph{NAACL-HLT}}. \bibinfo{pages}{5070--5083}.
\newblock


\bibitem[Z. and et~al.(2021a)]%
        {li2021implicit}
\bibfield{author}{\bibinfo{person}{Li Z.} {and} \bibinfo{person}{et al.}} \bibinfo{year}{2021}\natexlab{a}.
\newblock \showarticletitle{Implicit Representations of Meaning in Neural Language Models}. In \bibinfo{booktitle}{\emph{ACL-IJCNLP}}. \bibinfo{pages}{1813--1827}.
\newblock


\bibitem[Z. and et~al.(2022)]%
        {li2022automatingcodereviewactivities}
\bibfield{author}{\bibinfo{person}{Li Z.} {and} \bibinfo{person}{et al.}} \bibinfo{year}{2022}\natexlab{}.
\newblock \bibinfo{title}{Automating Code Review Activities by Large-Scale Pre-training}.
\newblock
\newblock
\showeprint{2203.09095 [cs.CL]}


\bibitem[Z. and et~al.(2020)]%
        {wu2020perturbed}
\bibfield{author}{\bibinfo{person}{Wu Z.} {and} \bibinfo{person}{et al.}} \bibinfo{year}{2020}\natexlab{}.
\newblock \showarticletitle{Perturbed Masking: Parameter-free Probing for Analyzing and Interpreting BERT}. In \bibinfo{booktitle}{\emph{ACL}}. \bibinfo{pages}{4166--4176}.
\newblock


\bibitem[Z. and et~al.(2021b)]%
        {wu2021explainingexplanationsbertempirical}
\bibfield{author}{\bibinfo{person}{Wu Z.} {and} \bibinfo{person}{et al.}} \bibinfo{year}{2021}\natexlab{b}.
\newblock \bibinfo{title}{On Explaining Your Explanations of BERT: An Empirical Study with Sequence Classification}.
\newblock
\newblock
\showeprint{2101.00196}


\end{thebibliography}

\end{document}